\documentclass[letterpaper, 10 pt, conference]{ieeeconf}
\IEEEoverridecommandlockouts

\usepackage{cite}
\usepackage{amsmath,amssymb,amsfonts}
\usepackage{algorithmic}
\usepackage{graphicx}
\usepackage{textcomp}
\usepackage{algorithm}
\usepackage{xcolor}
\usepackage{booktabs}
\usepackage{multirow}
\usepackage{fancyhdr}

\def\BibTeX{{\rm B\kern-.05em{\sc i\kern-.025em b}\kern-.08em
    T\kern-.1667em\lower.7ex\hbox{E}\kern-.125emX}}

\begin{document}

\pagestyle{fancy}
\fancyhf{}
\fancyfoot[C]{\footnotesize Preprint Version, Submitted to IEE RAL on 27th April 2025}
\addtolength{\headsep}{10pt}  
\renewcommand{\headrulewidth}{0pt}

\title{\LARGE \bf RICE: Reactive Interaction Controller for Cluttered Canopy Environment}

\author{Nidhi Homey Parayil$^{1}$, Thierry Peynot$^{2}$ and Chris Lehnert$^{3}$
\thanks{All authors are with the QUT Centre of Robotics, Queensland University of Technology, Brisbane QLD 4001, Australia (e-mail: nidhihomey.parayil@hdr.qut.edu.au). We acknowledge use of ChatGPT for minor text editing and grammar. This work has been submitted to the IEEE for possible publication. Copyright may be transferred without notice, after which this version may no longer be accessible.}}

\maketitle

\begin{abstract}

Robotic navigation in dense, cluttered environments such as agricultural canopies presents significant challenges due to physical and visual occlusion caused by leaves and branches. Traditional vision-based or model-dependent approaches often fail in these settings, where physical interaction without damaging foliage and branches is necessary to reach a target. We present a novel reactive controller that enables safe navigation for a robotic arm in a contact-rich, cluttered, deformable environment using end-effector position and real-time tactile feedback. Our proposed framework's interaction strategy is based on a trade-off between minimizing disturbance by maneuvering around obstacles and pushing through them to move towards the target. We show that over 35 trials in 3 experimental plant setups with an occluded target, the proposed controller successfully reached the target in all trials without breaking any branch and outperformed the state-of-the-art model-free controller in robustness and adaptability. This work lays the foundation for safe, adaptive interaction in cluttered, contact-rich deformable environments, enabling future agricultural tasks such as pruning and harvesting in plant canopies.

\end{abstract}



\section{Introduction}

Robots struggle to operate in an agricultural environment due to dense and unstructured clutter, such as overlapping leaves and branches~\cite{Legun2021Robot-ready:Robotics}. This clutter creates both physical obstructions, which require robots to interact with or navigate around obstacles, and visual occlusions, which hinder perception and path planning toward targets like fruits. When navigating cluttered environments, there are generally three possible strategies: pushing through obstacles, navigating around them, or adaptively combining both~\cite{Dogar2012AUncertainty}. However, most robotic systems are limited to the first two options~\cite{liang2021dexterous}, and obstacle avoidance is the most common approach in agricultural applications~\cite{silwal2022bumblebee}. As a result, recent developments in agricultural automation emphasize the selection of obstacle-free paths for tasks such as harvesting or pruning to minimize clutter interactions, thus limiting the application of robotic manipulators in agriculture~\cite{jin2024robotic}. This paper addresses the challenge of physical interaction and navigation to effectively reach a given target in cluttered agricultural settings.

   \begin{figure}[t]
      \centering
      \includegraphics[width=0.46\textwidth]{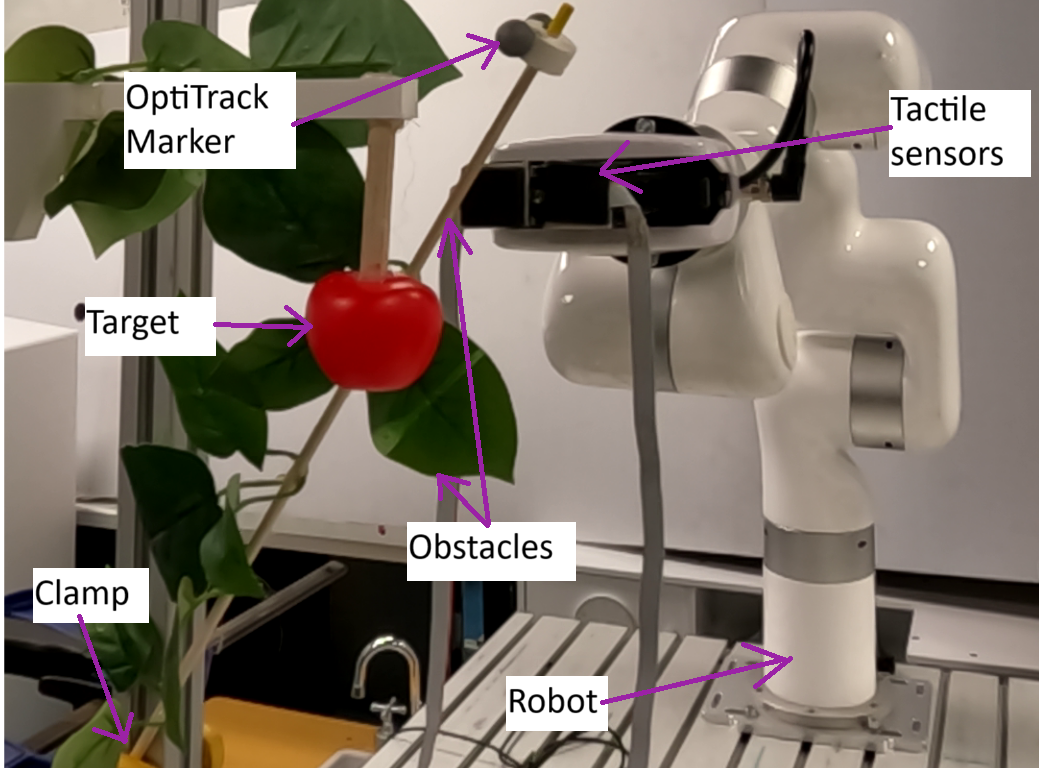}
      \vspace{-10pt}
      \caption{Experimental setup: In a cluttered canopy, the robot navigates toward a given target, encountering branches and leaves as obstacles.  OptiTrack markers measure branch disturbance. The robot adaptively pushes through or moves around obstacles to minimize disturbance using tactile feedback.}
      \label{fig:intro}
      \vspace{-20pt}
   \end{figure}

 The primary challenge for the deployment of robots within plant environments is generating motion that avoids damaging delicate structures like stems and branches~\cite{silwal2022bumblebee}. This demands control strategies that regulate both trajectory and applied force, often guided by visual or spatial information~\cite{Siciliano1999RobotControl}. In a cluttered environment, occlusions hinder perception, making it difficult to generate reliable trajectories.  Hybrid controllers are used for physical interaction by switching between position and force control based on task-specific criteria~\cite{IskandarHybridMotions}. However, clutter often causes frequent physical contact, requiring \emph{simultaneous} control of the path and force profiles~\cite{Siciliano1999RobotControl}.  During contact, the robot’s dynamics become coupled with those of the environment, complicating control and increasing reliance on accurate environmental models.

How do people handle visual occlusion when harvesting fruit among leaves and branches? While vision provides valuable spatial cues, it is often incomplete in dense foliage, where targets are partially or fully obscured. In such cases, humans naturally rely on tactile feedback to sense resistance. Leaves and branches are deformable, i.e. their shape and position change under contact, making the environment highly dynamic and difficult to model precisely. As we move, we feel how leaves and branches shift or bend, adjusting our motion to avoid excessive force or plant damage. In essence, we perform tactile servoing~\cite{suresh2024neuralfeels}. Without accurate dynamic models, it is hard to predict how the environment will respond to interaction. To handle such rapid and unpredictable changes, a controller must react in real time. Since classical planners struggle to accommodate fast, dynamic interactions~\cite{Frank2009Real-worldObstacles}, we use a reactive controller for this work.

To address the challenges posed by cluttered canopies, we propose a controller architecture called the Reactive Interaction Controller for Cluttered Canopy Environments (RICE). The key contributions of this paper are the following. We introduce a hierarchical, model-free control architecture that enables safe navigation in a contact-rich, cluttered, deformable environment using end-effector position and real-time tactile feedback to reach a target without damaging foliage and branches. We propose an optimal motion strategy that adaptively balances between pushing through to reach the target and maneuvering around obstacles to minimize interaction forces. To support quantitative analysis, we design a plant deformation measurement setup using a motion capture system, enabling capture of environmental interactions. Finally, we validate our approach quantitatively in custom-built, trackable mock plant environments across 30 distinct trials, as well as qualitatively in denser foliage with 5 trials, demonstrating superior robustness and adaptability compared to state-of-the-art model-free controllers.






\section{Prior work}
Navigating in a cluttered environment can involve various strategies depending on the task, but a common goal in agricultural tasks is to minimize damage to the robot or the system. This includes avoiding damage to stems, leaves, and trellis wires. Since free space is limited, the robot must often push through the environment to create a path. Motion planners designed for physical interaction \cite{Frank2014LearningPlanning, Grothe2022ST-RRT:Space-Time, Rodriguez2006PlanningEnvironments} are typically slow because they incorporate kinodynamic constraints. As a result, they fail to react quickly to unexpected obstacles in cluttered environments.  Reactive controllers make short horizon decisions based on real-time environment interaction feedback, enabling adaptation to changing dynamics in unpredictable environments \cite{Gieselmann2021Planning-augmentedLearning}. These systems often combine two control layers running at different frequencies \cite{Missura2021Fast-ReplanningA}: a high-level controller that responds to environmental changes and a low-level controller that manages the robot’s dynamics. Haviland \textit{et al.}~\cite{Haviland2021NEO:Manipulators} developed a reactive controller that uses quadratic programming on the robot model to optimize manipulability at the high level and employs Resolved Rate Motion Control (RRMC) at the low level. Their controller achieves faster performance than classical planners, but it does not support physical interaction because it does not use the model of the environment. 

Interaction control in cluttered environments involves both contact and non-contact phases, requiring the controller to regulate both position and force depending on the task context. Force-based controllers like admittance and impedance maintain desired force but only function when in contact \cite{Siciliano1999RobotControl}. Position-based controllers can push obstacles but focus on trajectory, instead of force. Hybrid controllers use position control in non-contact situations, and switch to force control once contact begins or when forces exceed a threshold \cite{Rhee2023HybridEnvironment}. Tasks, like surface cleaning, apply force control along one axis and position along another \cite{IskandarHybridMotions}. Iskandar \textit{et al.} developed a method to switch between force and position control independently on each axis~\cite{iskandar2024intrinsic, IskandarHybridMotions}. Wang \textit{et. al} used an approximate model of environment dynamics combined with a fuzzy logic-based hybrid controller to handle uncertainty~\cite{Wang2021HybridControl}. Building such models is difficult in cluttered, deformable environments, limiting their use in real-world agricultural settings.

Model Predictive Controllers (MPCs) are popular for interaction tasks because they can integrate multiple objectives and constraints effectively \cite{Gold2023ModelTasks}. However, this approach requires a model of the environment to define its constraints. In an agricultural environment, controlling interaction force precisely is often impractical, as a force suitable for one branch may damage another. One study in agriculture combined MPC with skin force sensors to navigate clutter made of leaves and logs \cite{Killpack2016ModelClutter}. The method demonstrated the robot pushing through compliant objects like leaves while avoiding rigid ones like logs.  It does not extend to interactions with elements like branches, which present additional challenges due to their deformability and fragility, and may require the robot to selectively push or maneuver around them.

An alternative approach in agriculture for navigating clutter involves manipulating the environment. One method uses a zig-zag pattern with multi-directional pushes to break contact with unripe strawberries, allowing access to hidden ripe ones \cite{Xiong2020AnClusters}. This strategy, though effective for fruit, struggles with leaves and stems, which have different dynamic properties. Lehnert et al. proposed moving the robot to improve visibility when occlusion occurs in cluttered environments \cite{lehnert20193d}. While this addresses visual occlusion, it does not solve the challenge of physical obstruction. Other methods plan around uncertainty by choosing paths with fewer obstacles or rearranging objects \cite{Dogar2012AUncertainty, Lee2021TreeClutter}. For crop monitoring, robotic arms have also been used to gently move leaves aside to reveal fruit \cite{yao2024safe}. Although manipulating obstacles can reduce occlusion, clearing all foliage can be time-consuming. 

Learning-based control techniques are gaining popularity for robots interacting with deformable objects. Most work focuses on 2D deformable objects such as cloth and ropes \cite{Erickson2020AssistiveRobotics, Matas2018Sim-to-RealManipulation}. Some approaches have addressed table-top clutter, such as a robot reaching a target hidden within soft objects like foam balls \cite{liang2021dexterous}. These environments are typically constrained to fixed surfaces with limited depth variation, and interaction occurs primarily in 2D. Agricultural environments are less structured and involve more complex, three-dimensional interactions with deformable elements like leaves and branches. Medical robotics has explored complex, unstructured, 3D deformable environments. For example, researchers have guided needles through soft tissue using demonstrations, inverse reinforcement learning, and accurate simulation models built from patient scans \cite{Tagliabue2020SoftSurgery}. While these techniques are accurate, they rely on extensive, high-fidelity data such as CT/MRI scans and anatomical models. Agricultural environments lack such data availability. Plant geometry and material properties vary widely across species, growth stages, and even environmental conditions. Although some work has explored plant simulation \cite{deng2024gazebo}, current tools and datasets remain insufficient to support detailed models for learning-based models in agriculture.

Navigating a robot through cluttered agricultural environments with physical interaction calls for a model-free, reactive controller capable of regulating position and force trajectory. The controller must make real-time decisions based on local tactile feedback to minimise environmental damage while effectively maneuvering around deformable plant structures. We propose a reactive control method that combines end-effector position with tactile force feedback, enabling a strategy that balances pushing through to reach the target and maneuvering around obstacles to minimal disturbance in dense clutter.

\section{RICE: Reactive Interaction controller for Cluttered canopy Environment}
In cluttered agricultural environments, humans instinctively move towards the target by choosing a path that optimally balances distance and environmental resistance. We navigate around stems, which are narrow and rigid, and push through compliant leaves. These conditions require a continuous trade-off between pushing through and moving around obstacles, rather than relying on a single optimal strategy. We exploit the fact that this trade-off is correlated with the spatial coverage of each object on the tactile sensors.  Trellis wires typically cover minimal area with most resistance, branches cover larger areas but can deform, and leaves may cover the entire sensor array yet allow compliant interaction. Drawing on this insight, we design our objective function to:
\begin{itemize} \item prioritize moving directly towards the objects when the robot detects no contact or uniform, low-resistance contact (e.g., from leaves),
\item maneuver around partial obstructions (e.g., stems) where excessive force could lead to plant damage.
\end{itemize}
We do not address interactions with large branches that completely occlude the sensor surface, as we consider these cases beyond the scope of this work. Our focus is to reach a target for tasks such as harvesting or pruning tasks involving plants like tomatoes, berry bushes and grapevines, where such large obstructions are uncommon.

   \begin{figure}
      \centering
      \vspace{2mm}
      \includegraphics[width =.48\textwidth]{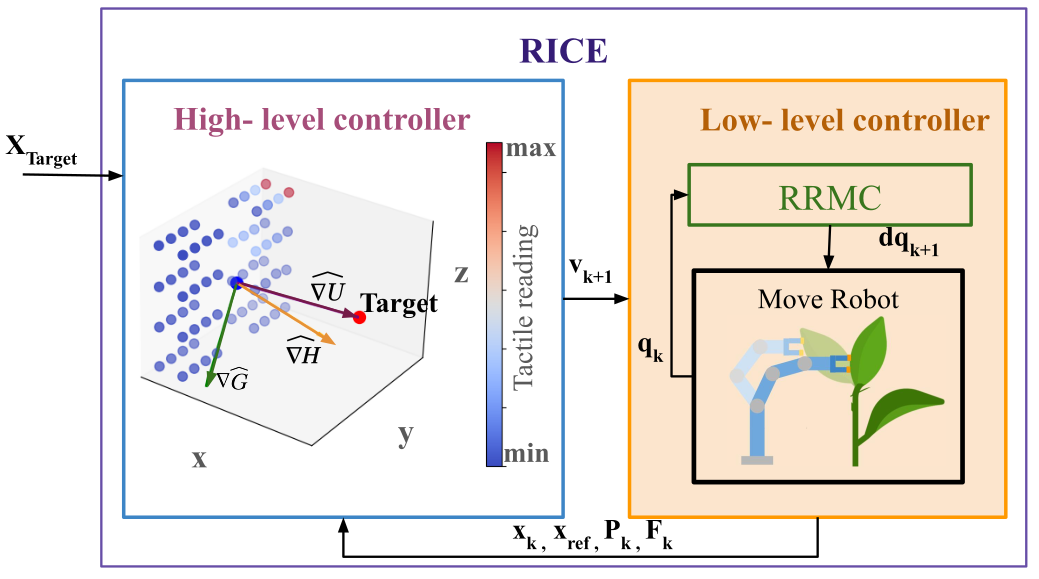}
       \vspace{-20pt}
      \caption{ Overall controller architecture showing a hierarchical scheme. The high-level controller uses end-effector position (\(\mathbf{x}_k\)), tactile forces (\(\mathbf{F}_k\)), and taxel positions (\(\mathbf{P}_k\)) to compute a velocity command toward the target (\(\mathbf{x}_{Target}\)). The low-level controller is a Resolved Rate Motion Controller (RRMC), which computes joint velocities from desired end-effector velocities using joint-state feedback.}
      \label{fig:flowchart}
      \vspace{-20pt}
   \end{figure}
\subsection{Sensor configuration}
   \begin{figure*}[t]
   \vspace{2mm}
      \centering
      \includegraphics[width =.98\textwidth]{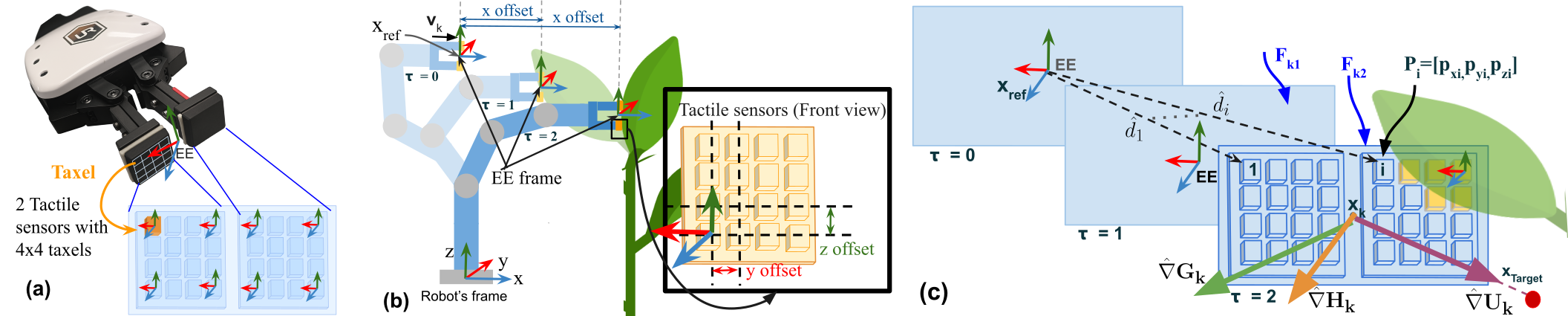}
       \vspace{-10pt}
    \caption{Left: Each fingertip is equipped with a \(4 \times 4\) tactile sensor array (\(n = 4\)), providing 16 taxels per sensor. EE represents the end effector's reference frame. Centre: Side view of the robot interacting with a leaf during one high-level control step, with the low-level controller running from \(\tau = 0\) to \(\tau = 2\) (\(j = 2\)). Forces along $\mathbf{x_{EE}}$ and $\mathbf{z_{EE}}$ are measured at each \(\tau\), while $\mathbf{x_{EE}}$ axis variation is estimated across steps. Right: The tactile sensors respond to contact with the leaf (yellow taxels).  RICE optimization computes motion direction \(\hat{\nabla} H\) as a weighted sum of the target reach cost \(\hat{\nabla} U\)  and interaction force cost \(\hat{\nabla} G\). \(\hat{d_i}\) is the direction vector from reference position to $i^{th}$ taxel and \( \mathbf{p_i}\) is the corresponding taxel position.}
      \label{fig:opt}
      \vspace{-15pt}
   \end{figure*} 
To sense interaction forces in cluttered environments, our system employs two tactile arrays, each mounted on a finger of the robot’s gripper as shown in Fig. \ref{fig:opt}(a). Each array features an \( n \times n \) taxel grid, yielding \( 2n^2 \) force measurements per time step of the low-level controller. Each taxel reports a three-axis force vector. We denote the end effector frame $EE$ as \(\mathbf{x_{\text{EE}}}, \mathbf{y_{\text{EE}}}, \mathbf{z_{\text{EE}}}\), for forward, lateral, and vertical to the gripper respectively. Given the planar placement of the arrays with known offsets in the \( \mathbf{y_{EE}} \) and \(  \mathbf{z_{EE}} \) directions, the system captures \( 2n^2 \) measurements in the end effector's frame at each instance of the low-level control cycle. To obtain force variation along the \( \mathbf{x_{EE}} \) axis, we record both tactile values and taxel positions across consecutive time steps of the low-level controller within a single high-level control cycle, as shown in Fig.~\ref{fig:opt}(b). Although skin-like tactile coverage across the entire manipulator would be ideal for operating in cluttered scenes, we limit sensing to the gripper tips due to hardware and cost constraints.
\subsection{Controller Architecture}
RICE is a hierarchical control architecture comprising a high-level, optimization-based controller operating at frequency \( f_{\text{H}} \) and a low-level controller running at a higher frequency \( f_{\text{L}} \). Their respective time steps are \( T = 1/f_{\text{H}} \) and \( \tau = 1/f_{\text{L}} \), and \( k \in \mathbb{N} \) is used for indexing high-level steps. The frequency ratio \( j = f_{\text{L}} / f_{\text{H}} \in \mathbb{N} \) indicates that the low-level controller executes \( j \) times between \( k-1 \) and \( k \). At each high-level step \( k \), the system collects \( j \) tactile frames from the low-level controller, denoted as \( \{ \mathbf{F}_{k_1}, \dots, \mathbf{F}_{k_j} \} \). Each frame \( \mathbf{F}_{k_m} \in \mathbb{R}^{N \times 3} \) contains 3D force vectors \(\mathbf{f} = [f_x, f_y,f_z] \in \mathbb{R}^3\) from \( N = 2n^2 \) taxels:
\[
\mathbf{F}_{k_m} = 
\begin{bmatrix}
f_{x1} & f_{y1} &f_{z1} \\
 & \vdots \\
f_{xN} & f_{yN} &f_{zN}
\end{bmatrix}, \quad m = 1, \dots, j
\]
To aggregate tactile data over the high-level interval, these frames are concatenated horizontally \(
\mathbf{F}_k = 
\begin{bmatrix}
\mathbf{F}_{k_1} & \cdots & \mathbf{F}_{k_is}
\end{bmatrix}^T
\in \mathbb{R}^{s \times 3}, \quad \text{where } s = N \cdot j
\). Corresponding taxel positions are represented by \( \mathbf{P}_k =[\mathbf{P}_{k_1}  \cdots  \mathbf{P}_{k_is}]^T \in \mathbb{R}^{s \times 3} \), with each column \( \mathbf{p}_i = [p_{xi}, p_{yi}, p_{zi}] \in \mathbb{R}^3\) indicating the spatial coordinates of the respective taxel. The high-level controller utilizes the end-effector position  \( \mathbf{x}_k  \in \mathbb{R}^3\), reference end effector position (at $\tau=0$ time frame) \( \mathbf{x}_{ref}  \in \mathbb{R}^3\),  target position \( \mathbf{x}_{Target}  \in \mathbb{R}^3\)  taxel positions \( \mathbf{P}_k \), and tactile feedback \( \mathbf{F}_k \) to compute an optimal end-effector velocity toward the target. The low-level controller translates this velocity into joint velocity\ \( \dot{\mathbf{q}}_k \in \mathbb{R}^b\) commands using joint position \( {\mathbf{q}}_k \in \mathbb{R}^b\) feedback where $b$ is the degree of freedom of the robot.
\subsection{High-level controller}
RICE computes the optimal velocity vector \( \mathbf{v}_{k+1} \in \mathbb{R}^3 \) at each high-level time step \( T = k \) by balancing two objectives: reaching the target and minimizing interaction forces. The objective function \( H_k \in \mathbb{R}^s \) is defined as a weighted combination of the target reach cost \( U(\mathbf{x}_k, \mathbf{x}_{\text{Target}}) \in \mathbb{R} \) and the force cost \( G(\mathbf{x}_k, \mathbf{P}_k, \mathbf{F}_k, \mathbf{x}_{\text{ref}}) \in \mathbb{R}^s \), scaled by scalar hyperparameters \( w_x, w_f \in \mathbb{R}^+ \):
\begin{equation}
{H_k} = w_x U(\mathbf{x}_k, \mathbf{x}_{\text{Target}}) + w_f G(\mathbf{x}_k, \mathbf{P}_k, \mathbf{F}_k, \mathbf{x}_{\text{ref}}).
\label{eq:dirvec}
\end{equation}
The optimal motion is given by the gradient descent of the objective function and its gradient is given by:
\begin{equation}
\nabla \mathbf{H_k} = w_x \, \hat{\nabla} \mathbf{U(\mathbf{x}_k}, \mathbf{x}_{\text{Target}}) + w_f \, \hat{\nabla}\mathbf{G(\mathbf{x}_k, \mathbf{P}_k, \mathbf{F}_k, \mathbf{x}_{\text{ref}}}),
\end{equation}
where \( \widehat{\nabla} \) denotes a normalized gradient to account for differences in units and scale.
Here, \( w_x \) emphasizes progress toward the goal (especially in free space), while \( w_f \) promotes deviation to reduce contact forces (e.g., when interacting with stems). This formulation allows the controller to adapt its trajectory in real time, balancing goal-directed motion with tactile-aware compliance (Fig. \ref{fig:opt}(c)).

\subsubsection{Target reach cost}
The gradient of the target reach cost \( \nabla \mathbf{U_k} \) is computed from a potential function inspired by potential field method \cite{khatib1986real}, where the robot is attracted toward a goal. The cost defined as the squared Euclidean distance between the current position \( \mathbf{x}_k \) and the target \( \mathbf{x}_{\text{Target}} \) given by  \(U_k = \| \mathbf{x}_{\text{Target}} - \mathbf{x}_k \|^2 \). The corresponding gradient is\(\nabla \mathbf{U_k} = -2 (\mathbf{x}_{\text{Target}} - \mathbf{x}_k) \). The normalised gradient is then: 
 \begin{equation}
  \hat{\nabla} \mathbf{U_k} = \frac{\nabla \mathbf{U_k}}{\| \nabla \mathbf{U_k} \|}.
 \end{equation}
\subsubsection{Interaction force cost}
The interaction force cost \( G \) is computed using tactile feedback \( \mathbf{F}_k \), corresponding taxel positions \( \mathbf{P}_k \), reference end-effector position \( \mathbf{x}_{\text{ref}} \), and current end-effector position \( \mathbf{x}_k \) during the \( k \)th step of the high-level controller. The idea of using an interaction force cost gradient across all axes to guide motion, minimizing contact forces in clutter, is inspired by the approach of Lehnert \textit{et al.}~\cite{lehnert20193d}, where visual gradients were used to move towards regions of higher visibility of fruits. We define direction vectors from the reference point \( \mathbf{x}_{\text{ref}} \) to each taxel position as:
 \begin{equation}
\hat{\mathbf{d}_k} = 
\begin{bmatrix}
\hat{\mathbf{d}_{k1}} & \cdots & \hat{\mathbf{d}_{ks}}
\end{bmatrix}^\top, \quad 
\hat{\mathbf{d}_{ki}} = \frac{\mathbf{p}_{ki} - \mathbf{x}_{\text{ref}}}{\| \mathbf{p}_{ki} - \mathbf{x}_{\text{ref}} \|} ,
\end{equation}
where \(i \in [0, s]\), \( \mathbf{p}_{ki} \in \mathbb{R}^3 \) denotes the Cartesian coordinates of the \( i \)th taxel, and \( \hat{\mathbf{D}_k} \in \mathbb{R}^{s \times 3} \) contains unit vectors pointing from \( \mathbf{x}_{\text{ref}} \) to each taxel. The force magnitude vector \( \mathbf{G}_k \in \mathbb{R}^s \) is computed as the Euclidean norm of each taxel’s 3D force:
\begin{equation}
{G}_k = 
\begin{bmatrix}
\| \mathbf{f}_1 \|_2 
\cdots 
\| \mathbf{f}_s \|_2
\end{bmatrix}^T .
\end{equation}

The reference magnitude is given by \( g_{\text{ref}} = \| \mathbf{f}_{\text{ref}} \|_2 \), where \( \mathbf{f}_{\text{ref}} \) is the average tactile force vector at \( \tau = 0 \). We define the interaction force cost deviation as:

\begin{equation}
\Delta \mathbf{G_k} = 
\begin{bmatrix}
\| \mathbf{f}_1||_2 - ||\mathbf{f}_{\text{ref}} \|_2 \\
\vdots \\
\| \mathbf{f}_s||_2 - ||\mathbf{f}_{\text{ref}} \|_2
\end{bmatrix}.
\end{equation}

Using the directional derivative approximation \(
\nabla \mathbf{G} \hat{\mathbf{d}_i}\approx \frac{g_i - g_{\text{ref}}}{\| \hat{\mathbf{d}_i} \|},
\) we model the deviation as, \(
\Delta \mathbf{G_k} = \hat{\mathbf{D}}_k \nabla \mathbf{G_k}.
\)
The spatial gradient \( \nabla \mathbf{G_k^*} \in \mathbb{R}^3 \) is then estimated using a least-squares solution:
\begin{equation}
\nabla \mathbf{G_k^*} = (\hat{\mathbf{D}}_k^\top \hat{\mathbf{D}}_k)^{-1} \hat{\mathbf{D}}_k^\top \Delta \mathbf{G_k}
\end{equation}
and the normalised gradient is given by \(\hat{\nabla} \mathbf{G_k}\).

\subsubsection{Optimal Velocity Calculation}
The optimal velocity vector \( \mathbf{v}_{k+1} \in \mathbb{R}^3 \) is computed using a normalised gradient descent method, based on the gradient vector \( \nabla \mathbf{H}_k \) and task-dependent step size \( \alpha \in \mathbb{R} \). The velocity is given by Eq.~\ref{eq:vel}:
\begin{equation}
\mathbf{v}_{k+1} = -\alpha \cdot \frac{\nabla\mathbf{H}_{k}}{|\nabla\mathbf{H}_{k}|}.
\label{eq:vel}
\end{equation}

\subsection{Low-level controller}
We used Resolved-Rate Motion Controller (RRMC)~\cite{rrmc}  for low-level motion control. The low-level controller is integrated with tactile sensing. RRMC maps the optimal end effector velocity \( \mathbf{v}_{k+1} \) calculated by the high-level controller into joint velocities \( \dot{\mathbf{q}}_k \) using the current joint configuration \( \mathbf{q}_k \) (see Fig~\ref{fig:flowchart}) using  Eq.~\ref{eq:rrmc}
\begin{equation}
\dot{\mathbf{q}}_{k+1} = J(\mathbf{q}_k)^+ \mathbf{v}_{k+1} + \mathbf{q}_k.
\label{eq:rrmc}
\end{equation}
RRMC uses the pseudoinverse of the Jacobian matrix \( J^+ \), offering improved robustness near singularities compared to traditional inverse kinematics approaches~\cite{Haviland2021NEO:Manipulators}. The high-level controller operates reactively, updating its velocity command based on real-time sensor feedback while allowing sufficient time for the low-level controller to execute each motion segment.
 \begin{figure*}[t]
  \centering
  \vspace{2mm}
  \includegraphics[width=.98\textwidth]{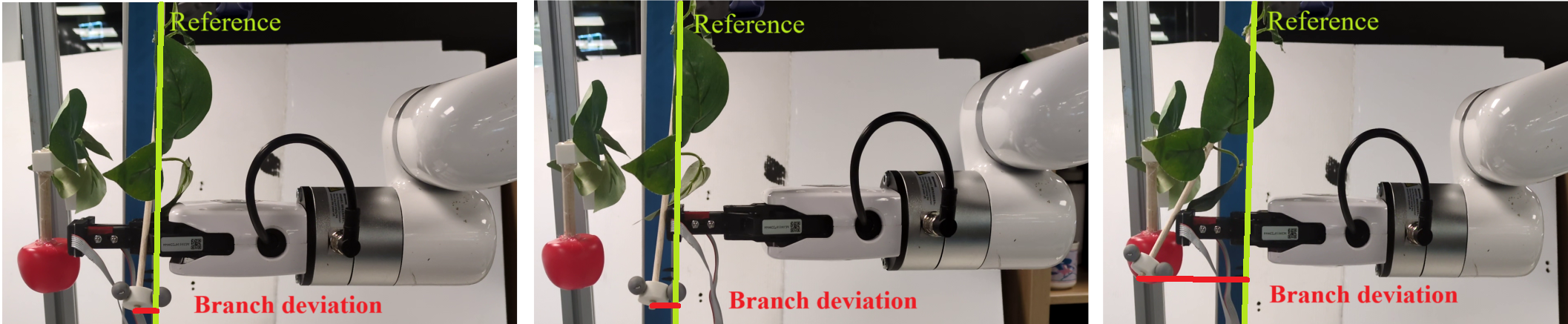}
  \vspace{-10pt}
  \caption{images comparing end states for the three controllers interacting with a single-branch experiment. RICE (left image) navigating around the obstacle to reach the target; the hybrid controller (centre image) pushing until the force threshold is exceeded, failing to reach the target; and the position controller (right image) pushing through the branch to reach the target.}
  \label{fig:comparison}
  \vspace{-15pt}
\end{figure*}

\subsection{Environment Assumptions}
To model deformation and interaction, we represent plant structures using a joint-based abstraction. External joints define attachments to parent structures (e.g., stem-to-branch) and allow global motion such as swaying. Each part is modelled as a chain of particles, where internal joints enable bending, with the number of particles determining length and flexibility. Breakage is defined as the inability to return to the original shape after exceeding displacement or torque limits. Environmental disturbance is quantified by measuring joint displacement during contact. We designed a mock plant to reflect the dynamics of external joints and the deformability of internal joints; further details are provided later on.

\section{System and Evaluation Framework}
We evaluated our controller’s ability to reach a target position compared to baseline and state-of-the-art controllers, using two types of custom-built trackable mock plant setups as shown in Fig.~\ref{fig:intro} and one unstructured artificial plant to depict realistic clutter. This section details the robotic system, custom-built trackable mock plant environment, evaluation metrics, and the procedure used to test and compare our proposed controller with baseline approaches.
\begin{table}[h]
 \vspace{-5pt}
\centering
\caption{Experimental Parameters Used in Evaluation}
 \vspace{-5pt}
\label{tab:experiment_parameters}
\begin{tabular}{ll}
\hline
\textbf{Parameter} & \textbf{Value} \\
\hline
Balsa wood cross-section & circular $\varnothing$(5/10/12)mm, square ($5\times5$)mm\\
Number of initial states & 5 \\
Obstacle  positions & 10 \\
Branch orientations (deg) & [60, 30, 0, -30, -60] \\
Step size (speed) & 0.01 m/s \\
Target positions & 15\\
\hline
\end{tabular}
 \vspace{-5pt}
\end{table}

\subsubsection{Robotic System} 
We use a UFactory XArm6 robot with a compact UFactory gripper~\cite{xarm6}. Two Xela uSPa44 tactile sensors~\cite{xela} are mounted on the gripper via a custom 3D-printed fixture (Fig.~\ref{fig:intro}). Each tactile sensor consists of $4\times 4$ taxel grids. The high-level controller runs at 50Hz, the low-level at 100Hz, and tactile data is sampled at 120Hz. The optimization algorithm takes an average of 1.7ms. The robot is stopped if it stops moving due to contact with a branch, significant deviation from the path, or collision with any part of the setup other than branches or leaves. Experiments are conducted on a workstation running Ubuntu 20.04 LTS with an NVIDIA RTX 3090 GPU. OptiTrack cameras and markers~\cite{optitrack} are used to track branch deviations.

\subsubsection{Custom trackable mock plant setup}
Trackable mock plants were built using balsa wood branches with artificial leaves, chosen for their realistic deformation and breakage properties. To reflect the circular cross-section of natural branches, we tested three different circular diameters, and to assess sensitivity to shape we included one square branch. All branches were securely mounted to frames to simulate attachment to a larger plant. OptiTrack markers were placed at each branch tip for measuring deviation (Fig.~\ref{fig:intro}). We tested four branch types with five initial joint configurations, ten obstacle placements, and fifteen target positions to span various contact scenarios and approach angles.Varying the branch orientation and position also resulted in different branch lengths, which in turn affected the stiffness and deformability. To ensure contact at the gripper tip where sensors are mounted, all branches were aligned parallel to it. Target positions were chosen to ensure reachability. Each trial recorded joint states and end-effector motion. Table \ref{tab:experiment_parameters} lists all the parameter variations used for our experiments. 
\subsubsection{Baseline Methods}
 To the best of our knowledge, no existing controller is designed for safe reactive navigation in contact-rich, cluttered environments like plant canopies, where the robot needs to push through foliage and branches, and perception is limited with no environmental model available. Therefore, we compare our method to two baseline controllers that use the two strategies commonly used in such settings. The first one is a Position Controller that drives directly towards the target, ignoring contact forces. This represents a simple and widely used approach in robotics where environmental resistance is minimal or contact is considered acceptable. The second is a state-of-the-art Hybrid Controller adapted from You et al.~\cite{you2022precision}, originally developed for guiding branches into a pruning blade using contact-aware motion. Their system applied force control in the y and z axes to reposition branches, but our setup differs as we use a planar end-effector designed to reach through foliage, with branches typically in front of the robot. We instead apply admittance control along the x-axis, the main direction of contact during forward motion, and retain position control elsewhere, where pushing is not needed. This shift in control axes supports our goal of gently displacing foliage to reach targets, rather than redirecting branches. The desired contact force (1N) was tuned experimentally for our specific plant setup to ensure safe interaction with thin branches. Corresponding admittance gains were tuned experimentally to optimize the controller performance with \( \text{diag}(M) = [0, 0, 0, 100, 0, 0] \) and \( \text{diag}(B) = [0, 0, 0, 50, 0, 0] \). All controllers use RRMC for low-level control, with a target speed of 0.01m/s to ensure consistency across evaluations.
\subsubsection{Evaluation Metrics}
To compare our controller with baseline methods, we used the following metrics:
\begin{itemize}
    \item \textit{Environmental Disturbance:} Defined as the maximum deviation of each branch from its initial position, with total disturbance in multi-branch setups given by the sum of individual deviations.
    \item \textit{Deviation from target:} Distance between the robot’s end-effector and the target at the end of the trial.
    \item \textit{No-Break Reach Rate:} Percentage of trials in which the robot reached the target without breaking any branch.
\end{itemize}
\subsubsection{Experimental Setups}
We conducted 5 experiments: \textbf{A. Parameter sweep:} A parameter sweep was conducted to select the optimal force weight ($w_f$), varying its value from 0.2 to 3 across 15 trials on a medium-thickness branch positioned vertically and obstructing one of the tactile sensors. All other parameters were kept constant. \textbf{B. Single branch setup:} We tested RICE and baseline controllers in 20 single-branch trials using four cross-sections and varying combinations of other parameters listed in Table~\ref{tab:experiment_parameters}. \textbf{C. Multi-branch setup:} A two-branch setup with varying cross-sections at a time was used to evaluate performance in a more dynamic and cluttered environment. Each controller was tested in 10 trials, varying combinations of other parameters listed in Table~\ref{tab:experiment_parameters}. \textbf{D. RICE repetitive trials:} To assess the repeatability and reliability of the RICE controller, we conducted five runs for each configuration in Experiment A and for five configurations in Experiment B. This resulted in 100 trials for the single-branch setup and 25 trials for the two-branch setup. \textbf{E. Artificial plant:} To evaluate performance in a more realistic setting, we conducted five tests on an artificial plant. Controller behavior was assessed visually, as OptiTrack was unsuitable due to clutter obstructing branch and leaf tracking. 
\section{Results}
Results and analysis for each experiment are detailed here.
\subsection{ Parameter Sweep}
   \begin{figure}[H]
   \vspace{-10pt}
      \centering
      \includegraphics[width=.48\textwidth]{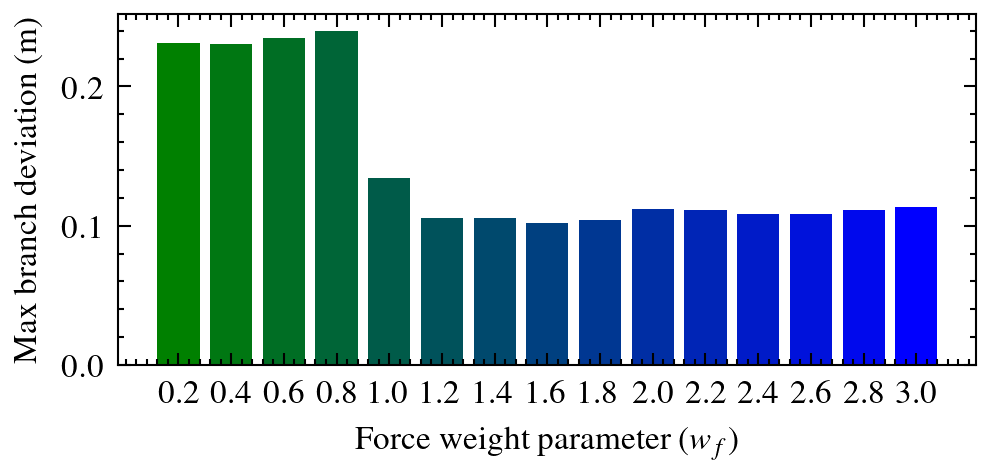}
      \vspace{-25pt}
      \caption{Maximum branch deviation for different force weight parameters. The robot exhibits motion around an obstacle for a force weight larger than 1.2. The variation in deviation of $w_f$ from 1.2 to 3 is due to the dynamic, uncertain nature of the branch.}
      \vspace{-10pt}
      \label{fig:param_sweep}
   \end{figure} 
Maximum branch deviation across different force weight values (\( w_f \)) is shown in Fig.~\ref{fig:param_sweep}. Since both the target cost and force cost gradients are normalized (\(\hat{\nabla}U_k\),\(\hat{\nabla}G_k\)), tuning only \( w_f \) is sufficient to balance goal-seeking and contact avoidance behavior. At low values (\( w_f < 1.2 \)), the controller tended to follow a straight-line trajectory, often pushing through obstacles and causing significant branch disturbance. In contrast, high values (\( w_f > 2 \)) led to repeated ineffective behavior: The robot would step back after contact but re-attempt the same trajectory, resulting in contact loops. Intermediate values (\( w_f = 1.4 \) to \( 1.8 \)) reduced disturbance but did not provide sufficient backward motion for effective recovery from sudden contacts. The best overall performance was achieved at \( w_f = 2 \), where the robot exhibited stable and adaptive behavior. Upon contact, it made small backward and lateral adjustments, allowing branches to recover from the disturbance and enabling it to re-approach the target from a different angle. This strategy reduced repeated contact and prevented motion loops critical in deformable, cluttered environments where prolonged interaction increases the risk of damage. Based on these findings, \( w_f = 2 \) was used for all subsequent experiments. Note that this value is specific to the current setup and may require retuning for different plant geometries or tasks.

\subsection{Single Branch}  
   \begin{figure}[H]
   \vspace{-15pt}
      \centering   
    \includegraphics[width = .45\textwidth]{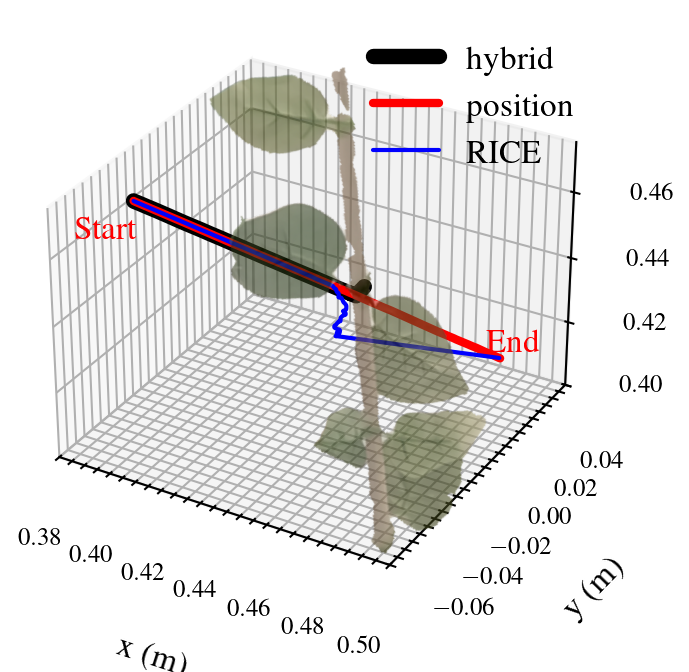}  
      \vspace{-10pt}
      \caption{Example of trajectories obtained with the three controllers interacting with a branch (experiment B). RICE (blue) successfully navigates around the branch to reach the target, the hybrid controller (black) stops upon contact, and the position controller (red) pushes through.}
      \label{fig:traj}
      \vspace{-5pt}
   \end{figure}
\begin{figure}[t]
  \centering   
  \vspace{2mm}
  \includegraphics[width = .48\textwidth]{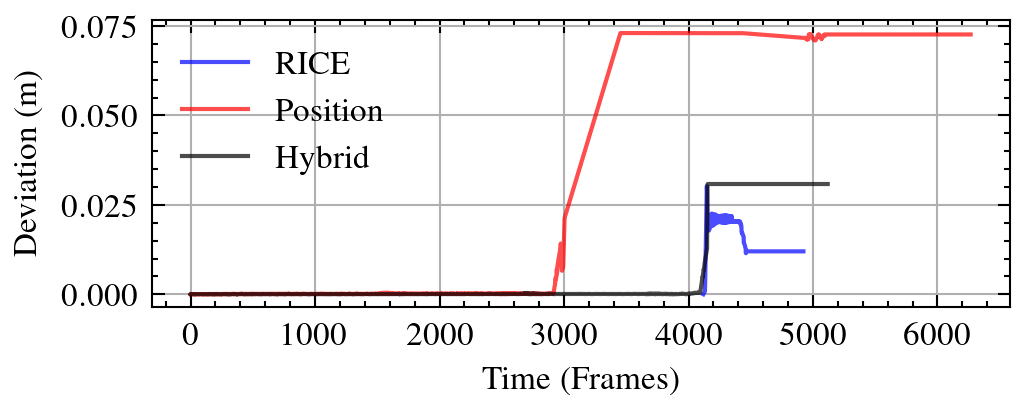}  
  \vspace{-28pt}
  \caption{Example of branch deviation recorded by the OptiTrack system. RICE (blue) causes minimal disturbance, the hybrid controller (black) pushes up to its 1\,N threshold, and the position controller (red) generates the highest disturbance.}
  \label{fig:dev}
  \vspace{-10pt}
\end{figure}

\begin{figure}[t]
      \centering
      \includegraphics[width = .48\textwidth]{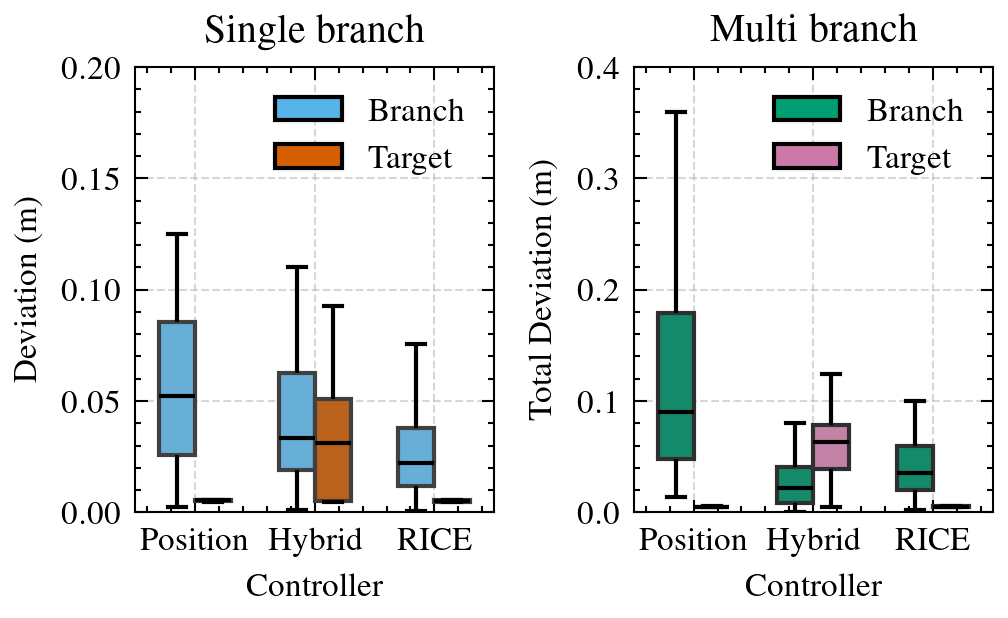}
    \vspace{-25pt}
      \caption{Branch deviation recorded from the OptiTrack and robot's deviation from target for 20 trials with a single branch (left) and 10 trials of multi-branches (right) for all three controllers  }
      \label{fig:box_plot}
      \vspace{-20pt}
   \end{figure}

  
Across 20 trials, the RICE controller successfully reached the target in every case without breaking any branches (Table~\ref{tab:controller_performance}). It achieved a median branch deviation of 22mm, adaptively navigating through foliage by pushing gently against leaf edges while avoiding stiffer midribs. The position-based controller also reached the target in all trials but caused the most disturbance, with five branch breaks and a median branch deviation of 52mm. Following a fixed path, it often forced through obstacles, even breaking thicker branches that did not deform easily. The hybrid controller reached the target in 9 out of 20 trials, causing three branch breaks. It advanced until reaching a contact force of 1N, then stopped, often getting stuck. Its median branch deviation was 33mm, with a median deviation from the target of 31.1mm. In one trial, where the motion involved movement along both the x and y axes to reach the target, the hybrid controller maintained contact along the x-axis but deviated significantly along the y-axis. In another, it broke a small branch clamped near the gripper due to its limited deformability. The results across all trials for the three controllers are summarized in Fig. \ref{fig:box_plot}, highlighting RICE’s consistent ability to minimize branch disturbance.. An example trajectory from a single test is shown in Fig.\ref{fig:traj}, where the position controller pushed through the branch, the hybrid controller stopped upon contact, and RICE maneuvered around it. The corresponding branch motion, recorded using the OptiTrack system, is shown in Fig.\ref{fig:dev}, indicating that RICE caused the least disturbance in this particular trial. The final end-effector and branch positions for this same test are illustrated in Fig.~\ref{fig:comparison}.

   
\subsection{Multi-Branch}
The RICE controller reached the target in all ten trials, with total branch deviation of 35mm (refer Fig \ref{fig:box_plot}) and no breakage, resulting in a 100\% No-Break Reach rate (see Table~\ref{tab:controller_performance}). It navigated around obstacles without causing damage, showing strong performance in complex, multi-branch setups. The hybrid controller reached the target in only one trial and failed to move past the first branch in the rest, resulting in a 10\% No-Break Reach rate. It had a median branch movement of 21.5mm and a deviation from target of 63mm, reflecting difficulty in adapting to multiple contacts. Its lower disturbance was due to limited interaction, unlike RICE, which contacted and passed both branches. The position-based controller followed a fixed path and broke branches in 9 out of 10 trials. It caused the most disturbance (median 90mm) but still reached the target in all cases. In Trial~1, it reached the target by bending the branches; in Trial~4, it broke both. This resulted in a 10\% No-Break Reach rate, indicating poor adaptability in constrained spaces. These results highlight the advantages of RICE`s responsive control strategies in cluttered plant environments, where adapting to contact and minimizing damage are critical.
\begin{table}
\vspace{2mm}
\caption{Controller performance for single and multi-branch setups}
 \vspace{-5pt}
\centering
\resizebox{0.5\textwidth}{!}{%
\begin{tabular}{llccc}
\toprule
\textbf{Experiment} & \textbf{Controller} & 
\begin{tabular}{c} No \\ breakage per trail \\  \end{tabular} & 
\begin{tabular}{c} Target \\ reached \\  \end{tabular} & 
\begin{tabular}{c} No-Break \\ Reach Rate \\  \end{tabular} \\
\midrule
\multirow{3}{*}{Single Branch} 
    & Hybrid         & 17/20 (broke 3 branches) & 9/20  & 6/20 (30\%)   \\
    & Position       & 15/20 (broke 5 branches) & \textbf{20/20} & 15/20 (75\%)  \\
    & RICE           & \textbf{20/20} (broke 0 branches) &\textbf{20/20} & \textbf{20/20} (100\%) \\
\midrule
\multirow{3}{*}{Multi Branch} 
    & Hybrid         & \textbf{10/10} (broke 0 branches) & 1/10   & 1/10 (10\%)   \\
    & Position       & 1/10 (broke 12 branches)  & \textbf{10/10}   & 1/10 (10\%)  \\
    & RICE           & \textbf{10/10} (broke 0 branches) & \textbf{10/10}   & \textbf{10/10} (100\%) \\
\bottomrule
\end{tabular}}
\label{tab:controller_performance}
 \vspace{-20pt}
\end{table}

\subsection{RICE repetitive trials}
To further assess the RICE controller, we ran five additional trials for Experiments B and C. The position and hybrid controllers were excluded due to their consistent behavior, either reaching the target with disturbance or failing. Across 100 single-branch trials, RICE reached the target every time without breaking any branches, with a median environmental deviation of 22mm. In 25 two-branch trials, it achieved a 100\% no-break reach rate (see Table~\ref{tab:controller_performance_d}), with a median total branch deviation of 36mm.
\begin{table}[h]
 \vspace{-5pt}
\caption{Performance of RICE Controller for repetitive trials}
 \vspace{-5pt}
\centering
\resizebox{0.5\textwidth}{!}{%
\begin{tabular}{lccc}
\toprule
\textbf{Setup} & 
\begin{tabular}{c} No breakage per trial\end{tabular} & 
\begin{tabular}{c} Target  Reached \end{tabular} & 
\begin{tabular}{c} No-Break Reach Rate \end{tabular} \\
\midrule
RICE (1 branch) & 100/100 & 100/100 & 100/100 \\
RICE (2 branches) & 25/25 (broke 0 branch) & 25/25 & 100/100\\
\bottomrule
\end{tabular}}
\label{tab:controller_performance_d}
 \vspace{-10pt}
\end{table}
\subsection{Artificial Plant}
We evaluated all three controllers in a highly cluttered environment using an artificial plant setup. Each controller was tested from the same initial configuration across five distinct target locations. The RICE controller successfully navigated around branches in all cases. The hybrid controller consistently halted upon contact, failing to progress toward the target. The position-based controller pushed through the branches, causing deformation or breakage. Representative behaviors and outcomes from these trials are shown in the supplementary media.
\section{Discussion}
We compared RICE with position and hybrid controllers, highlighting RICE's ability to adaptively push or maneuver around obstacles. This flexibility led to higher No-Break Reach Rates and lower disturbance, particularly in cluttered environments. The hybrid controller lacked consistent behavior to maneuver around obstacles, while the position controller rigidly pushed through all obstacles. Although RICE performed reliably in lab tests, future studies in an outdoor plant environment are recommended. Our experimental setup uses an industrial-grade robot not optimized for dense foliage, limiting maneuverability for non-tabletop configurations. Tactile sensors only at the gripper tip, expanding sensing, and adding vision could enhance environmental awareness and a better informed interaction strategy. In one of the trials for the single-branch experiment, temporary repeated contact and motion occurred as a highly deformable leaf midrib slid along the end-effector. This edge-case behavior resulted from the combined effects of obstacle configuration, trajectory direction, and sensor contact location. Although the robot initially displayed repeated motion behavior, it eventually reached the target, highlighting a transient interaction pattern rather than a failure, and highlighting the need for action randomness or belief-aware planning. The current objective function treats the robot as a point mass, ignoring link collisions, occasionally resulting in blocked paths. Finally, increasing the control and sensing frequency could improve adaptability to subtle environment dynamics. In some trials, where branches aligned with the robot's trajectory, the pushing strategy dragged branches toward the goal before eventually redirecting them, unlike human strategies, which typically displace obstacles away from the target. Future work will aim to investigate more intelligent pushing strategies. 
\section{Conclusion}
We introduced RICE, a reactive, model-free controller designed for safe navigation in cluttered, contact-rich environments such as agricultural canopies. To the best of our knowledge, RICE is the first method to combine real-time tactile feedback and end-effector position control in a hierarchical framework for adaptive navigation through deformable foliage without relying on an environmental model. This strategy enables motion that balances minimizing disturbance with safely pushing through obstacles to reach occluded targets. Our experiments across 30 trials in custom, trackable mock plant setups, and five additional trials in denser foliage demonstrated that RICE consistently reaches its target with lower disturbance and no damage, outperforming state-of-the-art model-free controllers in robustness and adaptability. These results highlight the potential of contact-aware, tactile-driven control for enabling safe and effective autonomous operation in unstructured agricultural scenarios. Future work will extend RICE to outdoor plant environments and explore the integration of additional sensing modalities to improve performance and robustness.
\bibliographystyle{IEEEtran}
\bibliography{IEEEabrv,ref}  
\vspace{12pt}
\end{document}